%% file: root.tex
\documentclass[letterpaper, 10 pt, conference]{ieeeconf}
\IEEEoverridecommandlockouts   
\usepackage{algorithm}
\usepackage{algorithmic}
\usepackage[bookmarks=true]{hyperref}
\usepackage{hyperref}
\usepackage{amsmath} 
\usepackage{amssymb}  
\usepackage{booktabs}
\usepackage{graphicx}
\usepackage[T1]{fontenc}
\usepackage[numbers,sort]{natbib}
\usepackage{colortbl}
\definecolor{Gray}{gray}{0.9}
\definecolor{White}{gray}{1}

\title{\LARGE \bf
DiffGen: Robot Demonstration Generation via Differentiable Physics Simulation, Differentiable Rendering, and Vision-Language Model
}

\author{Yang Jin$^{2*}$, Jun Lv$^{1*}$, Shuqiang Jiang$^{2,3\dag}$, Cewu Lu$^{1\dag}$
\thanks{*Equal contribution}
\thanks{$^{\dag}$Equal advising}
\thanks{$^{1}$Shanghai Jiao Tong University, $^{2}$University of Chinese Academy of Sciences, $^{3}$Institute of Computing Technology, Chinese Academy of Sciences}
}

\begin{document}

\maketitle
\thispagestyle{empty}
\pagestyle{empty}

\begin{abstract}

\input{tex/abs}

\end{abstract}

\section{Introduction}

\input{tex/intro}

\section{Related Work}

\input{tex/related}

\section{Technical Approach}

\input{tex/method}

\section{Experiments}

\input{tex/exp}

\section{Limitation and Future Work}

\input{tex/limit}

\section{Conclusion}

\input{tex/con}

\bibliographystyle{plainnat}
{\footnotesize
\bibliography{references}}

\end{document}

%% file: tex/abs.tex
Generating robot demonstrations through simulation is widely recognized as an effective way to scale up robot data. Previous work often trained reinforcement learning agents to generate expert policies, but this approach lacks sample efficiency. Recently, a line of work has attempted to generate robot demonstrations via differentiable simulation, which is promising but heavily relies on reward design, a labor-intensive process. In this paper, we propose DiffGen, a novel framework that integrates differentiable physics simulation, differentiable rendering, and a vision-language model to enable automatic and efficient generation of robot demonstrations.  Given a simulated robot manipulation scenario and a natural language instruction, DiffGen can generate realistic robot demonstrations by minimizing the distance between the embedding of the language instruction and the embedding of the simulated observation after manipulation. The embeddings are obtained from the vision-language model, and the optimization is achieved by calculating and descending gradients through the differentiable simulation, differentiable rendering, and vision-language model components, thereby accomplishing the specified task. Experiments demonstrate that with DiffGen, we could efficiently and effectively generate robot data with minimal human effort or training time.

%% file: tex/intro.tex
With the development of deep learning, significant success has been achieved in the fields of computer vision~\cite{he2016deep,he2017mask,kirillov2023segment}, language modeling~\cite{vaswani2017attention,devlin2018bert,achiam2023gpt,radford2021learning}, and context generation~\cite{goodfellow2020generative,song2020score,du2024learning}. Behind this progress, data has always played a key role. In recent years, researchers have showcased robots' capacity to acquire manipulation policies from demonstration data. By utilizing techniques like imitation learning~\cite{mandlekar2021matters,ho2016generative}, robots can effectively learn specific skills from these demonstrations. Scaling up the training data to encompass a variety of skills becomes a crucial next step in developing robot systems capable of performing diverse tasks in the physical world at the human level. Given the costliness of collecting real-world robot manipulation data, significant focus has been placed on generating such data in simulators~\cite{mandlekar2023mimicgen,wang2023gensim,wang2023robogen,katara2023gen2sim,xiang2020sapien,ma2023eureka}.

\begin{figure}[!t]
    \centering
    \includegraphics[width=\linewidth]{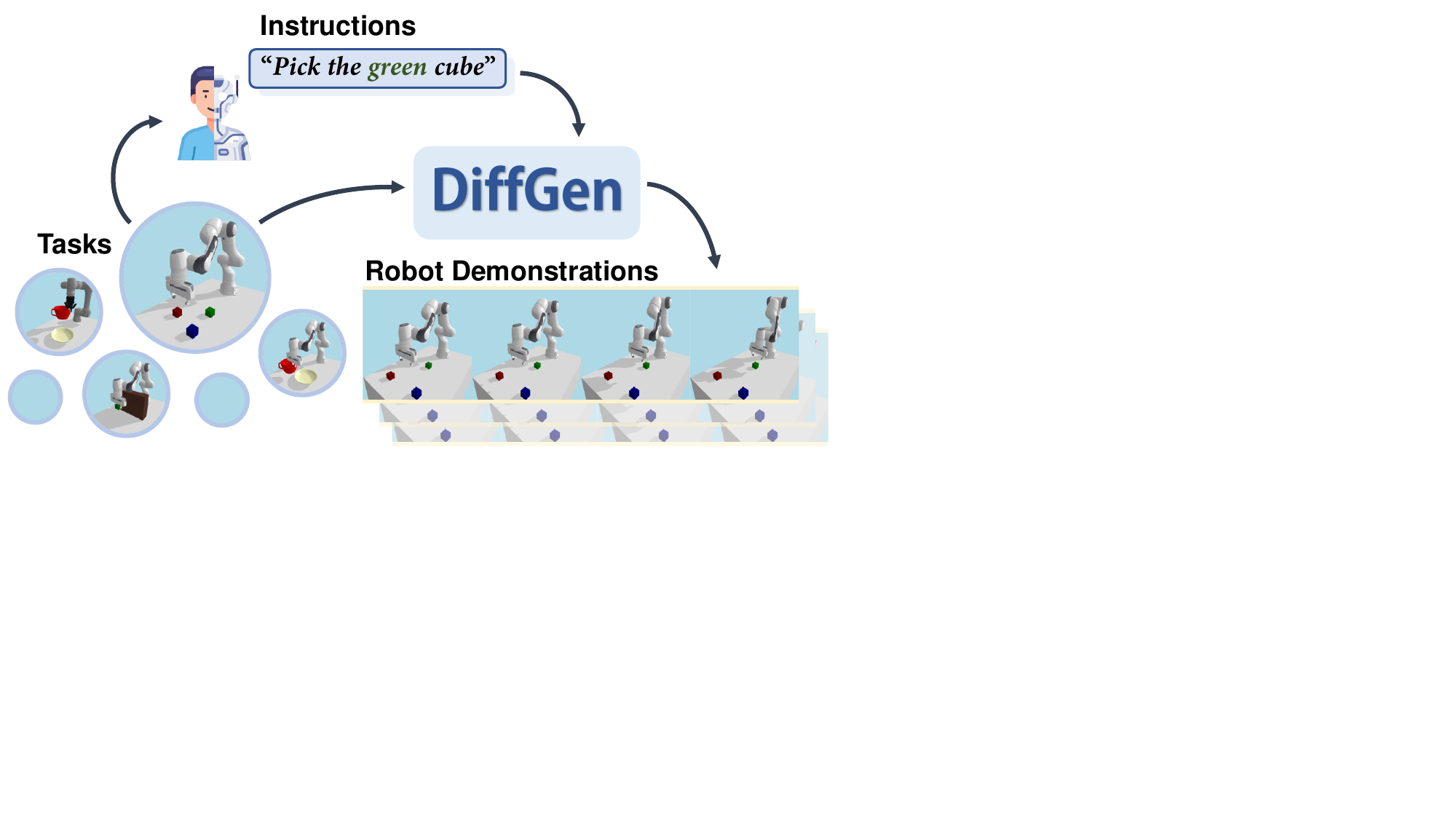}
    \caption{With DiffGen, we could automatically and efficiently generate robot demonstrations, given simulated task environments and text-based instructions.}
    \label{fig:teaser}
\end{figure}

There are two key steps in simulated robot manipulation data generation: scene construction and expert trajectory generation. With the advancement of generative artificial intelligence technique, automatic scene construction~\cite{wang2023gensim,wang2023robogen,katara2023gen2sim} can be achieved by utilizing large language models~\cite{achiam2023gpt} to generate scene descriptions and mesh generation models~\cite{liu2023zero} to create objects for the scene. In previous work, given a well-constructed task scene, the most popular approach is to train a policy via reinforcement learning to rollout manipulation trajectories~\cite{sontakke2024roboclip,ma2023eureka,wang2023robogen}. However, model-free reinforcement learning is criticized for its lack of sample efficiency and interpretability. To address this, a line of work is demonstrating the potential to utilize differentiable simulation to generate robot manipulation trajectories more efficiently and effectively~\cite{lin2022diffskill,lv2022sam,lv2022sagci,li2023dexdeform,Zhu2023DiffLfD,ren2023diffmimic}. To optimize the manipulation action, most approaches still rely on reward design by engineers, which is labor-intensive and hard to scale up. As shown in Fig.~\ref{fig:teaser}, in this paper, we are aiming to develop a novel system DiffGen to achieve automatic and efficient simulated robot demonstration generation by integrating differentiable physics simulation~\cite{de2018end,degrave2019differentiable,werling2021fast,howell2022dojo}, differentiable rendering~\cite{zhao2020physics,Li:2018:DMC} and vision-language pretrained model~\cite{ma2023liv,radford2021learning,sontakke2024roboclip}.

With differentiable physics simulation and rendering, visual observations can be generated to reflect the state change after robot manipulation. Previous works \cite{lv2022sam,gradsim,ma2021risp} have attempted to integrate them to address system or state identification problems. Inspired by recent progress in visual representation learning~\cite{nair2022r3m,ghosh2023reinforcement,bhateja2023robotic} and multi-modal learning~\cite{radford2021learning,sontakke2024roboclip,ma2023liv,huang2023voxposer}, our insight is that the integration of simulation, rendering, and cross-modal learning could effectively model the mapping between action and state, state and observation, and observation and instruction. By making all components differentiable, we can seamlessly connect each module and determine the mapping from task instructions to robot control signals, which will enable efficient robot demonstration generation.

Specifically, when given a manipulation scenario and a text-based task instruction, the system initiates the robot action sequence to control the robot in the differentiable physics simulation. Subsequently, by leveraging differentiable rendering, observation images are produced to depict the scene's state after the robot manipulation. A pretrained vision-language model is then utilized to evaluate the discrepancy between the image-based observation and the text-based task instruction, thereby serving as the objective function to guide the robot in completing the task. Thanks to the differentiability of the physics simulation, rendering, and the vision-language model, the system can directly explore feasible action sequences to execute the manipulation task in alignment with the task instruction, employing optimization-based methods such as gradient descent. Finally, simulated robot manipulation demonstration data is generated.

To evaluate the performance of our proposed system, we conducted several experiments. The results demonstrate that our system can generate simulated manipulation demonstrations for three different scenarios with distinct text-based task instructions. The proposed system exhibits better efficiency compared to reinforcement learning methods employing hand-crafted reward functions or reward functions generated by large language models.

The main contribution is summarized as follows:
\begin{itemize}
    \item We introduce a language-guided framework aimed at efficiently generating robot demonstrations.
    \item We integrate differentiable simulation, differentiable rendering, and a pretrained vision-language model to facilitate manipulation trajectory generation in simulation.
    \item We conduct experiments demonstrating the superior efficiency and reduced human effort of our new paradigm, compared with existing methods.
\end{itemize}

%% file: tex/related.tex
In this section, we provide a review of the literature related to the key components of our approach, including the generation of robot demonstration, integration of differentiable physics-based simulation and rendering, and reward learning. We also explain how our approach differs from previous works in these areas.

\subsection{Robot Demonstration Generation}
Robotics researchers have developed many different kinds of physics-based simulators~\cite{todorov2012mujoco,makoviychuk2021isaac,xiang2020sapien,brax2021github,werling2021fast} in the past and employed these simulators for robot demonstration generation. A burdensome aspect of using simulators is that it always requires engineers to build scenes for different tasks. However, with recent progress in generative artificial intelligence, such as large language models~\cite{achiam2023gpt} and mesh generation~\cite{liu2023zero,poole2022dreamfusion} from images or text, a line of work is aiming to build generative simulators to scale up robot demonstration generation.  For instance, \citet{wang2023robogen}, \citet{wang2023gensim}, and \citet{katara2023gen2sim} utilize large language models to automatically generate task descriptions and build the corresponding simulation environments. Additionally, \citet{lv2022sagci} constructs simulation environments from the real world and generates robot demonstrations within the simulation. \citet{mandlekar2023mimicgen} can generate large-scale robot data by augmenting human demonstrations. Our work focuses on how to generate robot demonstrations in specific task scenarios based on text-based instructions from humans or intelligent agents.

\subsection{Integration of Differentiable Physics-based Simulation and Rendering}
Physics simulations and renderers are essential tools in the domains of robotics and computer graphics. Recently, significant advancements have been achieved in the fields of differentiable physics simulation~\cite{de2018end,degrave2019differentiable,brax2021github,qiao2020scalable,werling2021fast,howell2022dojo,yang2023jade} and differentiable rendering~\cite{kato2020differentiable,liu2019soft,Li:2018:DMC,zhao2020physics}. \citet{gradsim} proposed a system that utilizes differentiable simulation and rendering to achieve system identification. In a similar vein, \citet{ma2021risp} leverage differentiable simulation and rendering for cross-domain parameter estimation from video. Additionally, \citet{lv2022sam} integrated differentiable simulation and rendering into robot manipulation learning. And \citet{Zhu2023DiffLfD} present contact-aware model-based learning from visual demonstration for robotic manipulation via differentiable simulation and rendering. While utilizing differentiable physics simulation to generate robot manipulation trajectories has become a popular research topic recently, many of these works still rely on hand-crafted reward functions, which is not optimal~\cite{lv2022sagci,lv2022sam,lin2022diffskill}.  In this paper, we propose combining differentiable physics simulation, differentiable rendering, and a vision-language model for specifying reward functions based on text-based task instructions.

\subsection{Reward Learning}
Reward learning is crucial for robot learning methods such as reinforcement learning~\cite{schulman2017proximal,haarnoja2018soft,fujimoto2018addressing}. A well-designed reward function can significantly improve performance and learning efficiency, whereas a poorly designed one may hinder progress. Traditionally, reward design has often relied on manual trial-and-error methods. However, recently, inferring reward functions from text-based instructions using large language models has gained popularity~\cite{yu2023language,ma2023eureka,wang2023robogen,huang2024diffvl}. For instance, \citet{ma2023eureka} has demonstrated the ability to achieve human-level reward design through coding language models, with algorithms capable of analyzing trained policy feedback to fine-tune reward functions. Another approach involves learning reward functions from demonstrations~\cite{bhateja2023robotic,ghosh2023reinforcement,ma2022vip,nair2022r3m,ho2016generative}. \citet{ma2022vip} presented a self-supervised pretrained visual representation, enabling the generation of reward functions for unseen robotic tasks based on this pretrained representation. To enable cross-modal reward learning~\cite{ma2023liv,sontakke2024roboclip,radford2021learning}, \citet{ma2023liv} introduced a unified objective for vision-language representation and reward learning from demonstrations with text annotations, training multi-modal representations to serve as reward functions across language and image domains. In this paper, we integrate the cross-modal reward learning method with differentiable physics simulation and rendering to generate robot demonstrations.

%% file: tex/method.tex
\label{sec:Method}

\begin{figure*}
    \centering
    \includegraphics[width=\linewidth]{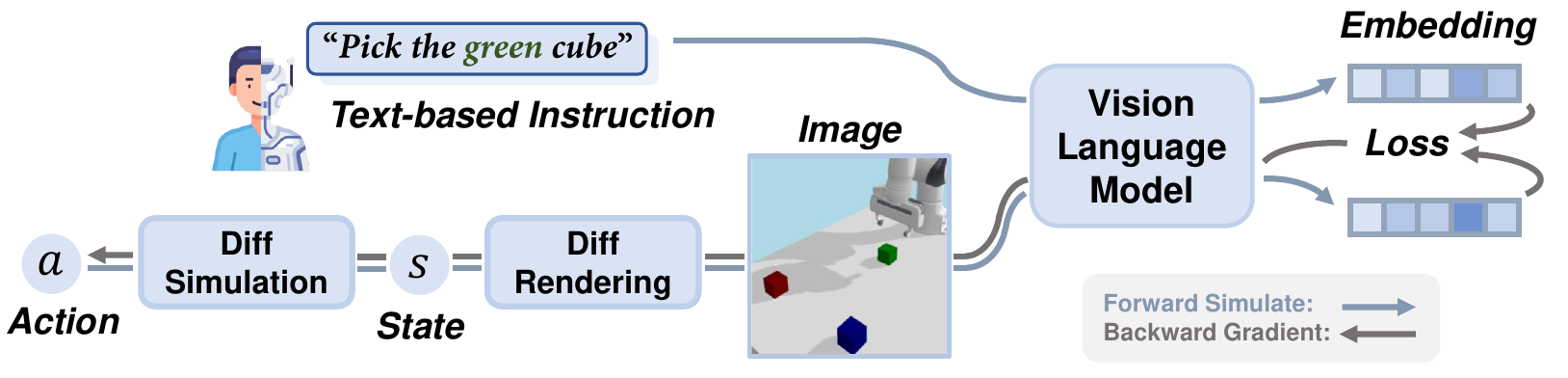}
    \caption{The overall pipeline of our proposed DiffGen. Our system first initiates action sequences, simulates the state changes, and renders visual observations after manipulation, via differentiable simulation and differentiable rendering. Then, a vision-language model is employed to measure the distance between the visual observations and the text-based instructions. Thanks to the differentiability of each component, feasible action sequences can be generated by gradient descent optimization.}
    \label{fig:overview}
\end{figure*}

\subsection{Overview} 
Given a robot manipulation task, where the task goal $\mathcal{G}$ is specified by a text-based instruction $l$, which can be generated by a language model~\cite{achiam2023gpt} or provided by a human, the problem addressed in this paper is robot demonstration generation. The objective is to find a sequence of actions $\{a_t\}_{t=0}^T$ that control the robot to modify the state $s_t$ to reach a final goal state $s^\mathcal{G}$ that satisfies the instruction $l$. Specifically, we aim to maximize the following objectives:
\begin{equation}
    \begin{gathered}
        \max_{a_t} \mathbb{E}\left[\sum_{t=0}^T \gamma^t \mathcal{R}(s_t, a_t;\mathcal{G})\right],\\
        \text{s.t. } s_{t+1} = \mathcal{T}(s_t, a_t),\ s_t\in\mathcal{S},\ a_t\in\mathcal{A},
    \end{gathered}
\end{equation}
where $\mathcal{S}$ denotes the state space, $\mathcal{A}$ denotes the action space, $\mathcal{T}:\mathcal{S}\times\mathcal{A}\to\mathcal{S}$ is the transition map, and $\mathcal{R}$ is the goal-conditioned reward function.

Our proposed approach, as shown in Fig.~\ref{fig:overview}, leverages recent advances in differentiable physics simulation, differentiable rendering, and vision-language pretraining to optimize an action sequence $\{a_t\}_{t=0}^T$ that achieves the goal $\mathcal{G}$. In the following sections, we will describe how to utilize the differentiable physics simulation and rendering to model the environment dynamics (Sec. \ref{sec:DiffSimRender}), how to achieve goal specification via a vision-language model (Sec. \ref{sec:clip}), and how to generate actions via an optimization-based method (Sec. \ref{sec:opt}).

\subsection{Differentiable Physics-based Simulation and Rendering}\label{sec:DiffSimRender}
With an action sequence $\{a_t\}_{t=0}^T$, a physics-based simulation $f(\cdot, \cdot)$ takes the current state $s_t$ and action $a_t$ to calculate the next state $s_{t+1}$ via a forward dynamic function:
\begin{equation}
    s_{t+1} = f(s_t, a_t).
\end{equation}
When the forward function is formulated in a differentiable manner~\cite{werling2021fast,yang2023jade}, the gradients of the next state with respect to the current state $\frac{\partial s_{t+1}}{\partial s_t}$ and with respect to the action $\frac{\partial s_{t+1}}{\partial a_t}$ can be obtained through back-propagation. 
\begin{equation}
\begin{aligned}
    \frac{\partial s_{t+1}}{\partial s_t} = \frac{\partial f(s_t, a_t)}{\partial s_t}, 
    \\
    \frac{\partial s_{t+1}}{\partial a_t} = \frac{\partial f(s_t, a_t)}{\partial a_t}.
\end{aligned}
\end{equation}
Previous work has optimized action sequences that transition the environment to the goal state via gradient descent~\cite{lv2022sagci}. However, the goal state is usually directly specified by a human or determined by a hand-crafted objective function, which is labor-intensive. In our proposed method, we aim to specify the goal state via a pretrained vision-language model. Therefore, our next step is to generate the visual observation $\mathcal{I} \in \mathbb{R}^{H \times W \times 3}$ using a renderer $g(\cdot)$, where $\mathcal{I}$ is an RGB image with height $H$ and width $W$:
\begin{equation}
    \mathcal{I}_t = g(s_t).
\end{equation}
In practice, the state $s_t$ comprises the transformation pose of each object $\mathcal{O}_i$ in the scene. We transform each object to its simulated pose to generate the visual observation that describes the current state. And when the renderer is also differentiable~\cite{Li:2018:DMC}, the gradient of each pixel on the image $\mathcal{I}$ with respect to the state $\frac{\partial \mathcal{I}}{\partial s_t}$ is also available:
\begin{equation}
    \frac{\partial \mathcal{I}}{\partial s_t} = \frac{\partial g(s_t)}{\partial s_t}.
\end{equation}
Now, the proposed system enables us to achieve goal specification in the visual observation space.

\subsection{Goal Specification via Vision-Language Model}\label{sec:clip}
Goal specification is a crucial step in robot demonstration generation, involving the assessment of whether the current state aligns with the provided task instruction. To achieve this, given a text-based instruction $l$, we utilize a vision-language model $h(\cdot)$, which is pretrained for value learning (for more details, please refer to LIV \cite{ma2023liv}), to encode the instruction:
\begin{equation}
    z^l = h(l),
\end{equation}
while the visual observation $\mathcal{I}$ could be also encoded via:
\begin{equation}
    z^\mathcal{I} = h(\mathcal{I}).
\end{equation}
When the action sequence transitions the environment to a state $s_t$ that fulfills the task instruction $l$, the distance between the embedding vector of the text-based instruction $z^l$ and the embedding vector of the visual observation $z^\mathcal{I}$ should be minimized. So the objective function $\mathcal{L}(\cdot,\cdot)$ under the setting of our proposed system is:
\begin{equation}
    \mathcal{L}(z^l, z^\mathcal{I}) = -\frac{z^l\cdot z^\mathcal{I}}{\|z^l\|\|z^\mathcal{I}\|}.
\end{equation}

\subsection{Robot Demonstration Generation via Optimization} \label{sec:opt}
To generate the robot demonstration, the system aims to find a robot action sequence $\{a_t\}_{t=0}^T$ that can change the environment state to $s^\mathcal{G}$, such that the rendered visual observation $\mathcal{I}$ has a similar embedding vector $z^\mathcal{I}$ to the embedding vector of the text-based task instruction $z^l$. Previous works typically train a policy network via model-free reinforcement learning with the assistance of a reward function, which suffers from poor sample efficiency. In our proposed system, thanks to the differentiability of the physics simulation, rendering, and the neural-network-based vision-language model, we can directly backpropagate the gradient and search for feasible action sequences that minimize the objective function $\mathcal{L}$ via gradient descent:
\begin{equation}
    \{a_t\}_{t=0}^T \leftarrow \{a_t\}_{t=0}^T - \lambda \frac{\partial \mathcal{L}}{\partial \{a_t\}_{t=0}^T},
\end{equation}
\begin{equation}
    \frac{\partial \mathcal{L}}{\partial \{a_t\}_{t=0}^T}=\frac{\partial \mathcal{L}}{\partial z^\mathcal{I}} \frac{\partial z^\mathcal{I}}{\partial \mathcal{I}} \frac{\partial \mathcal{I}}{\partial s_T} \frac{\partial s_T}{\partial \{a_t\}_{t=0}^T},
\end{equation}
where the $\lambda$ is the optimization step size. With this optimization-based demonstration generation, the system could generate robot data effectively and efficiently.

As mentioned in studies~\cite{ren2023diffmimic,brax2021github}, previous optimization methods based on differentiable physics simulation often suffer from gradient vanishing or explosion, likely to get stuck in local minima. This is because the gradients from the objectives have to be backpropagated through the entire sequence of state transitions, which may lead to error accumulation and gradient instability.

To address this issue, we design a novel optimization strategy to decompose the long-horizon trajectory into a series of short-horizon episodes. Each short episode is optimized individually, one by one, with a higher learning rate and lower optimization steps to enable fast exploration. The cumulative trajectory keeps expanding until the objective function reaches the extreme or the iteration number reaches the limit, at which point the process is automatically stopped.

%% file: tex/exp.tex
In this section, we introduce the implementation details of the proposed system and the experimental setup of this paper. We conduct experiments to evaluate the efficiency and performance of the proposed system.

\subsection{Implementation Details}

\subsubsection{Differentiable Physics Simulation}
In our experiments, we use the differentiable physics simulator NimblePhysics~\cite{werling2021fast} as our simulation backend. NimblePhysics provides Jacobians of the environment dynamics while supporting collision handling simultaneously. This is achieved by solving the Linear Complementarity Problems (LCPs). Most of the computationally intensive parts of the simulation are implemented in C++ for high performance. For more details, please refer to \cite{werling2021fast}.

\subsubsection{Differentiable Renderer}
We replace the default rendering module of NimblePhysics with a differentiable renderer, Redner~\cite{Li:2018:DMC}, to enable the end-to-end differentiability of the entire pipeline. We use the fast deferred rendering mode of Redner to ensure correct gradient estimation, fast backpropagation speed, and high-quality rendering results. For more details, please refer to \cite{Li:2018:DMC}.

The output state of NimblePhysics, which is represented in joint space, is converted to a list of transformation matrices to meet the input requirements of Redner. Those transformation matrices describe the absolute positions and orientations of all simulated links in Cartesian space. We implement this conversion in a differentiable manner to avoid breaking the gradient flow.

\subsubsection{Vision-Language Model}
The vision-language representation model used in our experiments is LIV~\cite{ma2023liv}, a CLIP-style model with ResNet-50~\cite{he2016deep} as its vision backbone and DistilBERT~\cite{sanh2019distilbert} as its language encoder. The model is initialized with weights from CLIP~\cite{radford2021learning} and pretrained on EpicKitchen~\cite{Damen2022RESCALING}, which is a large-scale first-person non-robotic video dataset, consisting of human daily activities in the kitchen, annotated with natural language descriptions. The pretrained models are also fine-tuned with robot data to reduce the domain gap. We fine-tune the vision encoder with a learning rate of $1e-5$ and a batch size of $64$ following the default setting in LIV. The language encoder is frozen during fine-tuning. The fine-tuned encoders are then used to extract features and construct rewards. We exclusively utilize a restricted scale of data around 400k. In Sec.~\ref{sec:generalizability}, we demonstrate that the model's performance is not overfitting on this dataset.

\subsubsection{Gradient-based Optimization}
To perform gradient-based optimization on the actions, we use the AdamW optimizer with a learning rate of $1e-2$ for most tasks. For the fixed-horizon strategy, the optimization steps and horizon lengths are decided according to the complexity of each task.

\subsection{Task Setup}

\begin{figure}[!h]
    \centering
    \begin{minipage}[c]{0.32\linewidth}
        \centering
        \includegraphics[width=\linewidth]{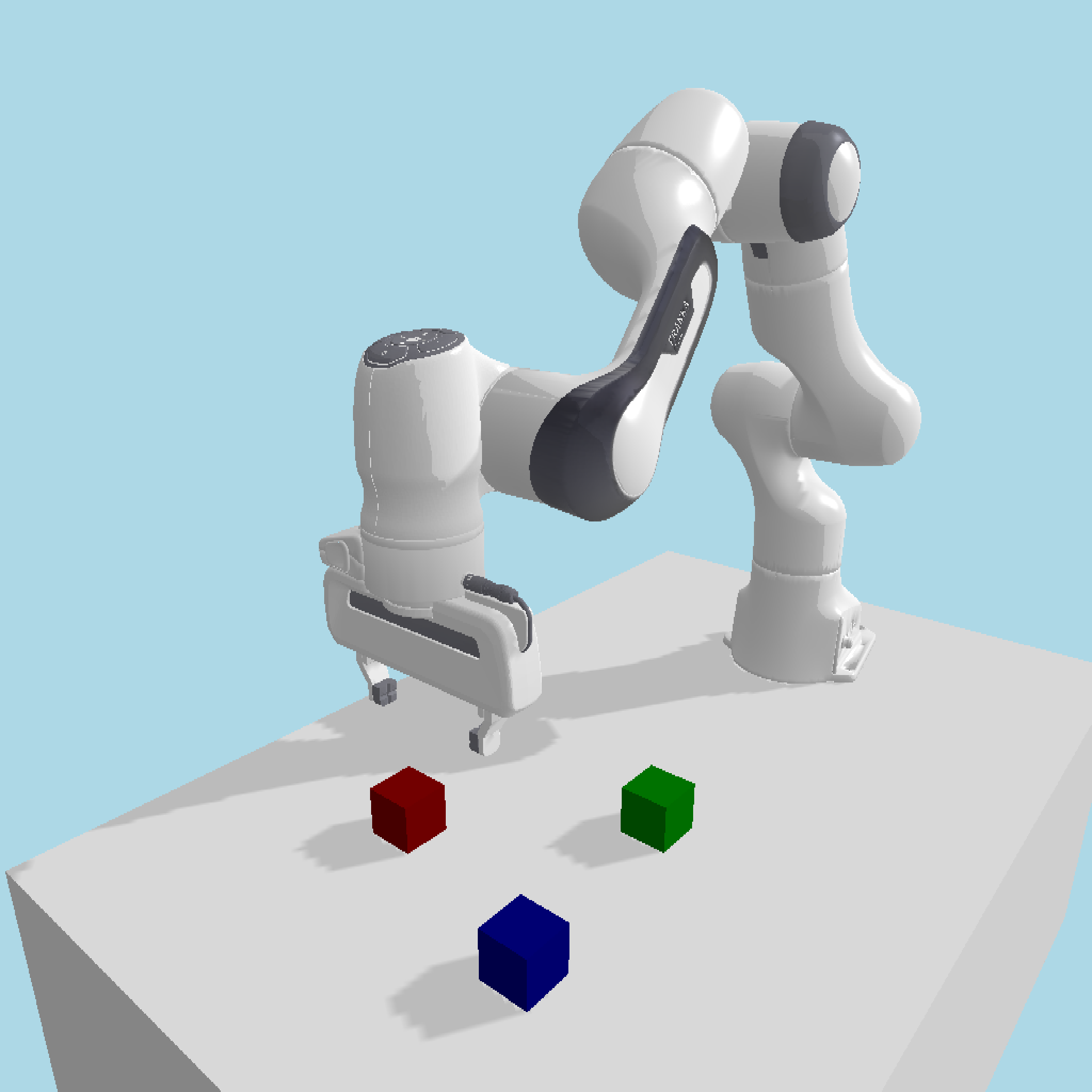}
        \small\textit{Cube-Selection}
    \end{minipage}
    \begin{minipage}[c]{0.32\linewidth}
        \centering
        \includegraphics[width=\linewidth]{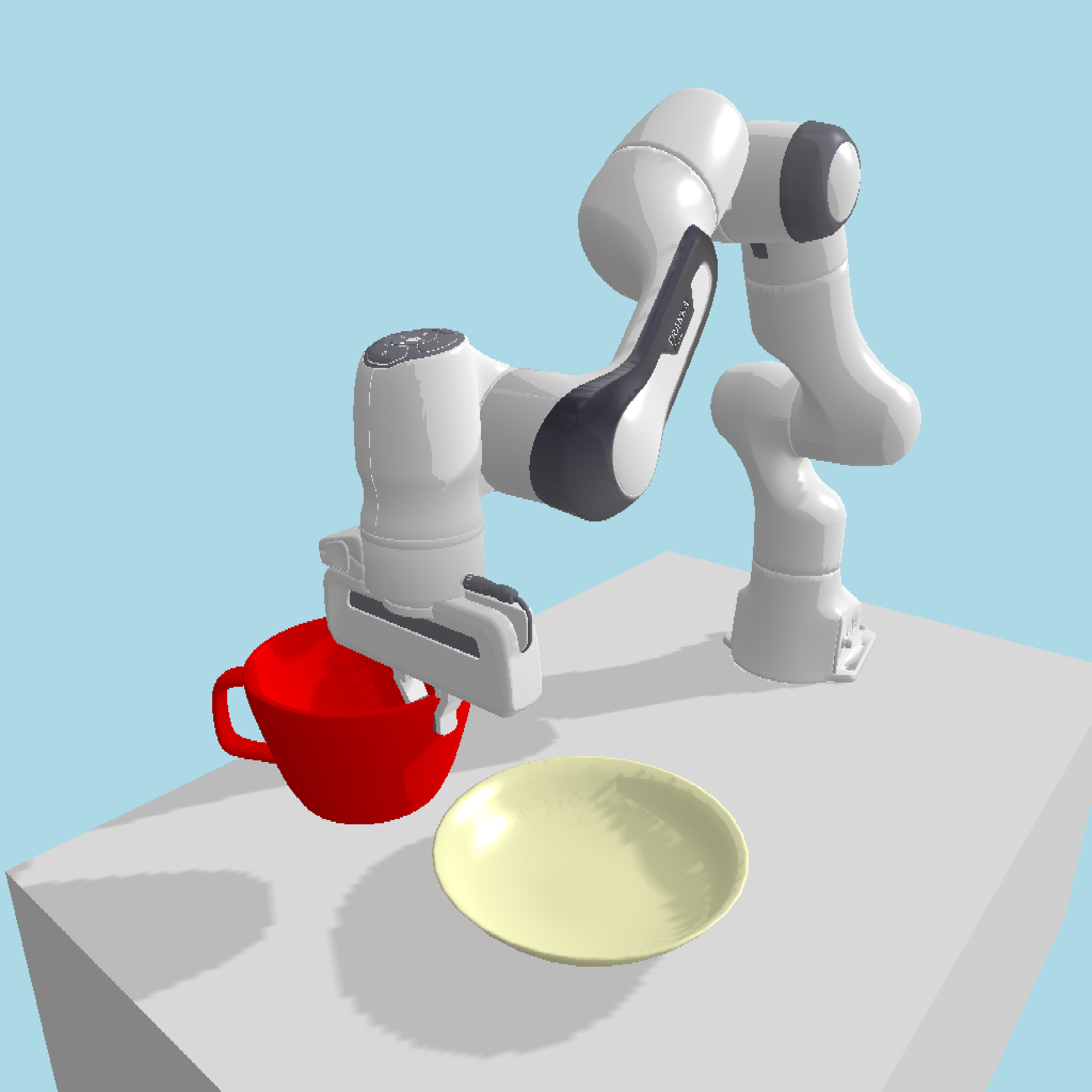}
        \small\textit{Cup-Placing}
    \end{minipage}
    \begin{minipage}[c]{0.32\linewidth}
        \centering
        \includegraphics[width=\linewidth]{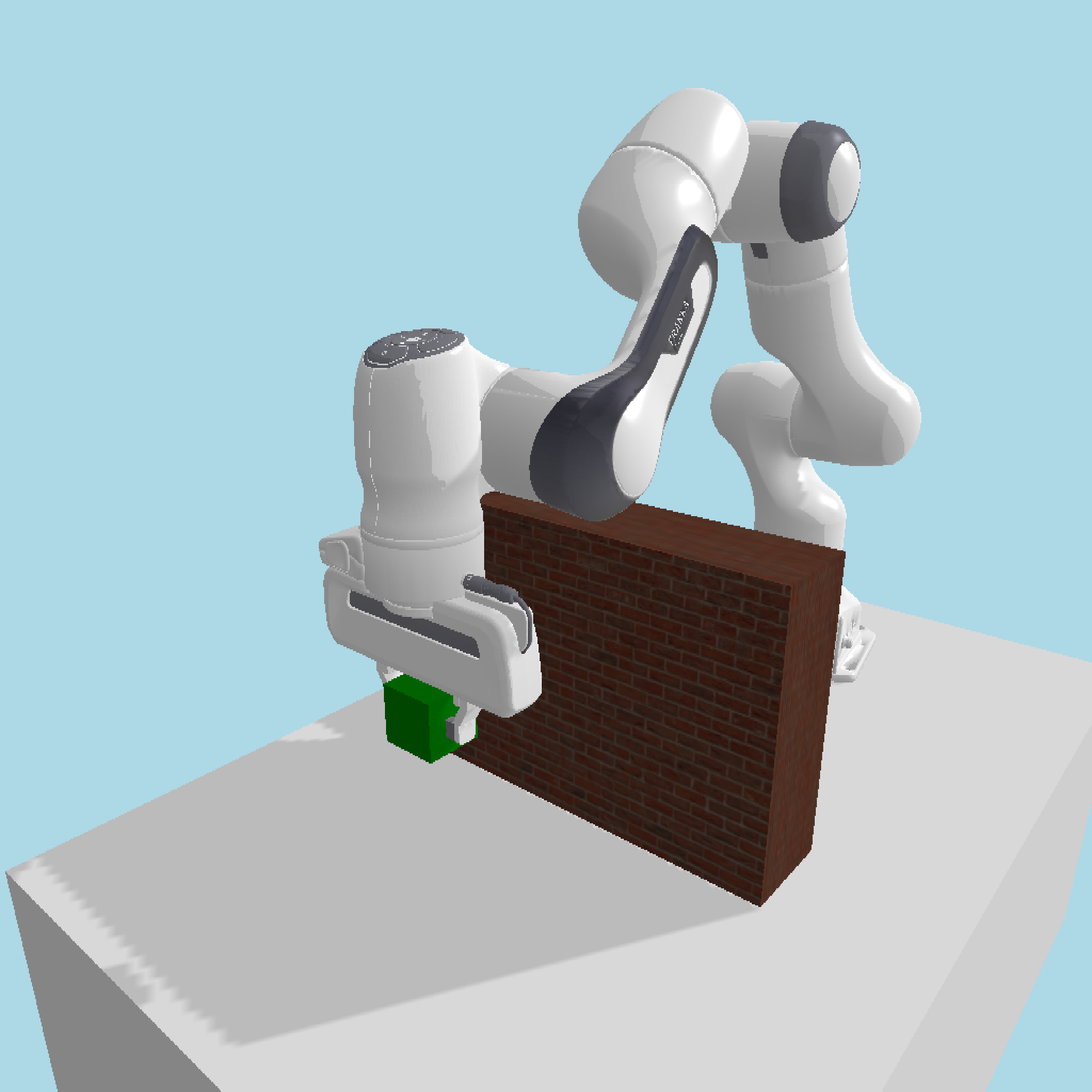}
        \small\textit{Obstacle-Crossing}
    \end{minipage}
    \caption{Visualization of the tasks, generated by the PyBullet renderer.} \centering
    \label{fig:TaskSetup}
\end{figure}

As shown in Fig.~\ref{fig:TaskSetup}, we design three tasks to demonstrate the performance of our method. All these tasks are evaluated in simulation on a Franka Emika Panda robot arm. The arm is equipped with a default gripper, placed on a square platform, and initialized with random end-effector positions. Though we use Franka Emika Panda in most cases, our method is general and can be applied to other robot embodiments, such as UR5e, as will be demonstrated in Sec. \ref{sec:generalizability}.

Specifically, these tasks are:

\subsubsection{Cube-Selection}
The job is to distinguish the cube in the target color and select it. Cubes of different colors (red, blue, green) are placed on the platform at random positions. Given a brief language instruction, such as \textit{"grasp the red cube"}, the robot is expected to move its gripper to the target cube. We consider the task successful if the gripper is within a certain distance above the target cube. The deviation threshold is set to $10cm$. To handle this task, we optimize the robot trajectory for $200$ steps with a fixed horizon of $50$ timesteps.

\subsubsection{Cup-Placing}
The job is to place a cup onto the dish. The cup is already held by the robot arm and the dish is placed on the platform at a random position. The robot is instructed to \textit{"put the cup on the dish"}. The task is considered successful if the cup is accurately put inside the dish in the final state, which is measured by the distance between the cup bottom and the dish surface. After scrutinizing the cup and dish at different relative positions, we set the threshold to $5cm$, since $5cm$ is much smaller than the radius of the dish, indicating the center of the cup bottom is inside the dish. The optimization process is set to $200$ steps and the horizon is set to $70$ timesteps in this case. Since it is a task that requires precise manipulation, we decrease the learning rate to $1e-3$ to ensure the stability of the optimization process.

\subsubsection{Obstacle-Crossing}
The job is to move the cube to the commanded position while avoiding the obstacle. The obstacle is implemented as a wall located between the initial and the target place. The size of the wall is $30cm \times 40cm \times 10cm$. The commands can be \textit{"move to the left side of the wall"}, \textit{"move to the right side of the wall"}, or \textit{"move to the back of the wall"}. The task is considered successful if the robot crosses the obstacle and reaches the target position. For example, if the command is \textit{"move to the back of the wall"} and the cube on the robot hand is delivered to the $40cm \times 30cm$ region behind the wall, we consider the task a success. We set the optimization steps to $300$ and the horizon length to $100$.

\subsection{Evaluating the Efficiency and Accuracy}

\subsubsection{Robot Demonstration Generation}

We show that our method is capable of generating high-quality robot demonstrations on all three tasks. To this end, we set up 100 randomly initialized environments for each task and apply our method to those environments. Each trajectory is optimized individually for a predefined number of steps with a fixed horizon. The optimal trajectory can be considered as either the one with the highest reward during the whole optimization process, denoted as \textit{Ours (best traj.)}; or the last trajectory, denoted as \textit{Ours (last traj.)}. The performance of our method is evaluated based on the collected trajectories, averaged over all environments. 

\begin{figure}[!htbp]
    \centering
    \begin{minipage}[c]{0.49\linewidth}
        \centering
        \includegraphics[width=\linewidth]{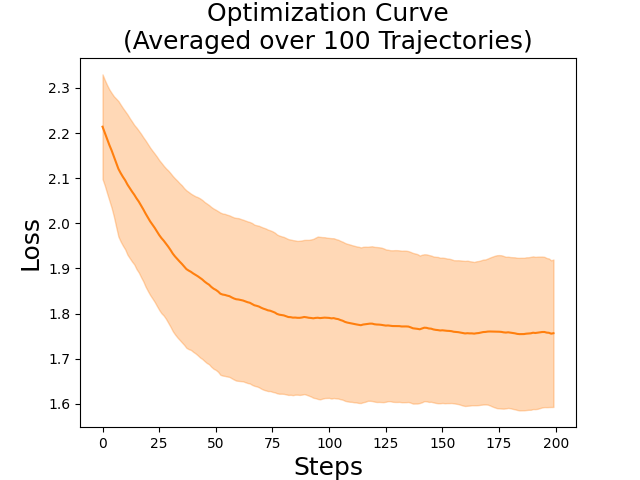}
    \end{minipage}
    \begin{minipage}[c]{0.49\linewidth}
        \centering
        \includegraphics[width=\linewidth]{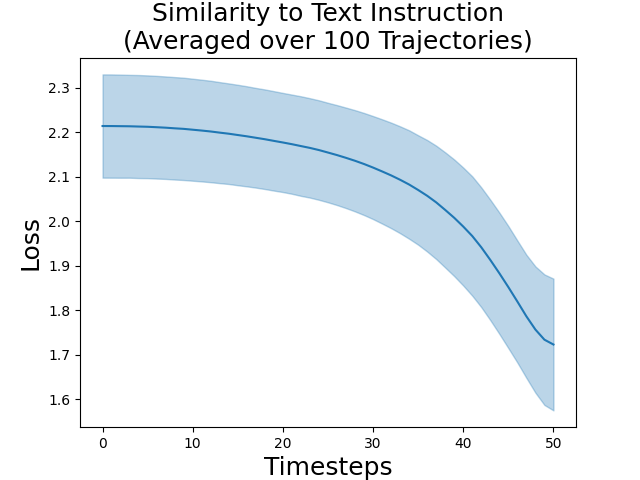}
    \end{minipage}
    \caption{Loss Curves on Cube-Selection Task} \centering
    \label{fig:ColorIdentification}
    \vspace{5pt}
    \centering
    \begin{minipage}[c]{0.49\linewidth}
        \centering
        \includegraphics[width=\linewidth]{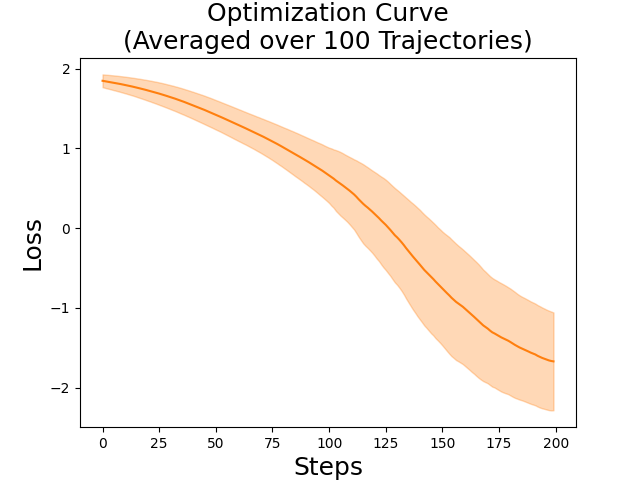}
    \end{minipage}
    \begin{minipage}[c]{0.49\linewidth}
        \centering
        \includegraphics[width=\linewidth]{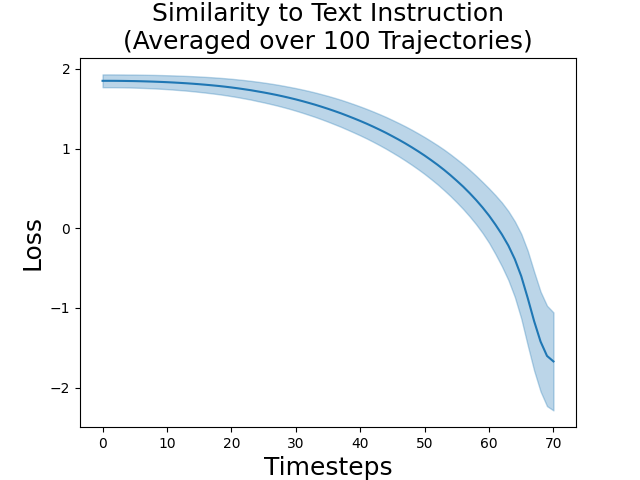}
    \end{minipage}
    \caption{Loss Curves on Cup-Placing Task} \centering
    \label{fig:CupPlacing}
    \vspace{5pt}
    \centering
    \begin{minipage}[c]{0.49\linewidth}
        \centering
        \includegraphics[width=\linewidth]{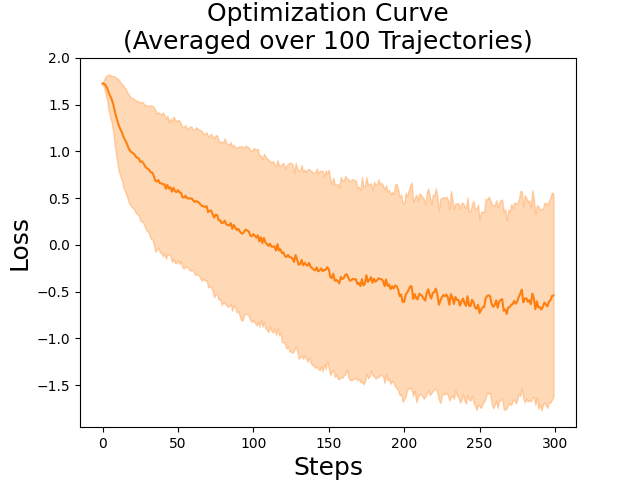}
    \end{minipage}
    \begin{minipage}[c]{0.49\linewidth}
        \centering
        \includegraphics[width=\linewidth]{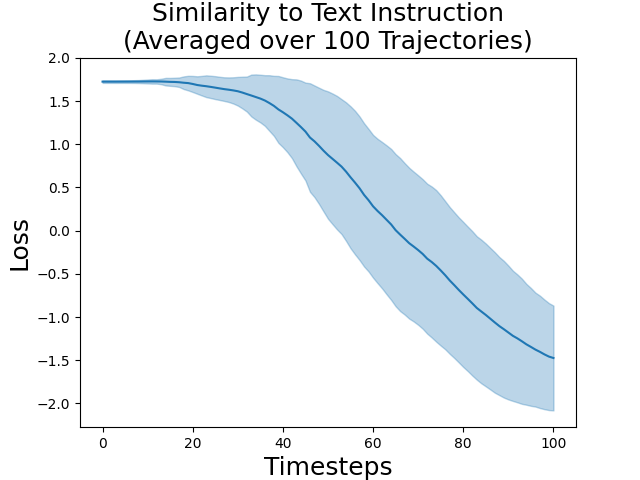}
    \end{minipage}
    \caption{Loss Curves on Obstacle-Crossing Task} \centering
    \label{fig:ObstacleCrossing}
\end{figure}

We present the loss curves of our method in Fig.~\ref{fig:ColorIdentification}, Fig.~\ref{fig:CupPlacing}, and Fig.~\ref{fig:ObstacleCrossing}. The left column shows the trajectory-level losses with respect to the optimization steps. As we can see, the loss decreases steadily and converges rapidly as the optimization process proceeds. The convergence is achieved within a few hundred steps, indicating the efficiency of our method. The right column shows the loss curves of those optimized trajectories. The loss can be regarded as the embedding distance between the current observation and the text instruction. As time goes by, the loss keeps decreasing, suggesting that the robot is approaching the goal. The robot stops at a state with relatively low loss. This indicates the potential of our method in generating smooth and accurate robot demonstrations.

\subsubsection{Comparison with Reinforcement Learning Baselines}
A popular way to generate robot demonstrations is to train a reinforcement learning agent with reward functions generated from vision-language models (VLMs) or large language models (LLMs). We choose two representative paradigms from past works as our baselines~\cite{ma2023eureka,sontakke2024roboclip}.

\begin{itemize}
    \item \textit{VLM-based Methods,}
    which train a PPO agent with a trajectory-level reward generated by some vision-language model, such as S3D~\cite{xie2018rethinking}. We follow the implementation in RoboCLIP~\cite{sontakke2024roboclip} to construct the reward as the similarity score between the task descriptor and the observed agent behaviors. We replace the original S3D model with fine-tuned LIV in our experiments to ensure a fair comparison. Apart from the cosine similarity, we also set up a baseline that uses the InfoNCE~\cite{oord2018representation} loss as the reward function. Those two baselines are denoted as \textit{PPO-CosSim} and \textit{PPO-InfoNCE}, respectively.

    \item \textit{LLM-based Methods,}
    which obtain the reward function from some large language models with code-writing capabilities, such as GPT-4~\cite{achiam2023gpt}. The entire environment of the task, along with natural language prompts, is fed into those LLMs to generate codes of reward functions, which are then embedded into the PPO training pipeline. We follow the prompt designs in Eureka~\cite{ma2023eureka} and keep the reward reflection approach proposed in that work. This baseline is denoted as \textit{PPO-LLM}.
\end{itemize}

We compare our method to the baselines in terms of the success rate, mean deviation of the final position from the target, and the total optimization steps. For both baselines, we first train the agents using the PPO algorithm. The trained policies are then evaluated on 100 trajectories to obtain an average performance. The results are shown in Table~\ref{tab:ColorIdentification}.

\begin{table}[!htbp]
    \caption{Quantitative Results on Cube-Selection Task} \centering
    \label{tab:ColorIdentification}
    \begin{tabular}{l c c c}
        \toprule
        Method & Success Rate & Deviation & Tot. Opt. Steps \\
        \midrule
        Ours (best traj.)       & 0.85 & 0.082 & 20k \\
        Ours (last traj.)       & 0.81 & 0.095 & 20k \\
        \midrule
        PPO-CosSim              & 0.06 & 0.446 & 10M \\
        PPO-InfoNCE             & 0.00 & 0.259 & 10M \\
        PPO-LLM (first iter.)   & 0.25 & 0.274 & 5M \\
        PPO-LLM (second iter.)  & 0.37 & 0.173 & 5M \\
        PPO-LLM (third iter.)   & 0.40 & 0.224 & 5M \\
        \bottomrule
    \end{tabular}
\end{table}

Our method outperforms all baselines and achieves a relatively high success rate with lower computation costs. Simply applying LIV-generated rewards to PPO training, as demonstrated in \textit{PPO-CosSim} and \textit{PPO-InfoNCE}, fails to learn a reasonable policy. 
It indicates that, by directly backpropagating the gradients of visual feedback to the action space, our method achieves higher sample efficiency compared to RL baselines.

\subsection{Evaluating the Generalizability}
\label{sec:generalizability}

Robot data is known as expensive to collect and hard to be reused. Human data, on the other hand, is easy to scale up while facing severe domain gaps. To bridge the gap between humans and robots, we use some robotic data to fine-tune the vision-language model after pretraining it on human data. We argue that the model is not overfitted to those robotic data and can be generalized to new embodiments or scenarios. We conduct the following experiments to demonstrate this:

\subsubsection{Zero-shot Goal Specification}

We conduct evaluations on the Obstacle-Crossing task. The vision-language model for reward specification is trained on the human data from EpicKitchen only, without any fine-tuning. We set the language instructions to be \textit{"move to the left side of the wall"} or \textit{"move to the right side of the wall"}, and expect the robot to be optimized to move left or right accordingly. The success is defined as the cube ending up with a positive horizontal displacement if commanded left, and negative if commanded right. This is challenging since the robot has never encountered any robotic data before, and can only infer the concept of "left" and "right" from human videos. The results are shown in Table~\ref{tab:ZeroShotGoalSpecification}, from which we can see that most of the time the robot can tell left from right and move accordingly, indicating the potential of our method in transferring human demonstrations to robot tasks.

\begin{table}[!htbp]
    \caption{Zero-shot Goal Specification on Obstacle-Crossing Task} \centering
    \label{tab:ZeroShotGoalSpecification}
    \begin{tabular}{l c c c}
        \toprule
        Command & Success Rate & Displacement & Final Position \\
        \midrule
        Move Left   & 0.978 & -0.205 & -0.217 \\
        Move Right  & 0.704 &  0.049 &  0.050 \\
        \bottomrule
    \end{tabular}
\end{table}

\subsubsection{Cross-Embodiment Transfer and Generalization}

To evaluate the performance of the proposed method on the unseen robotic embodiment, we conduct experiments on the Cup-Placing task. We first fine-tune the vision-language model with data from Panda Emika Franka and then evaluate its performance on a UR5e robot. The challenge in this experiment lies in the embodiment gap between those two robots. To successfully place the cup onto the dish, the robot needs to realize that the concepts of "put" and "on" are invariant across different embodiments and associate the language instructions with the consequences of its actions, rather than embodiment-specific behaviors. Apart from the cross-embodiment transfer experiment, we also evaluate whether the robot can learn from demonstrations of different scenarios and generalize to a new one. To this end, we change the color of the cup during evaluation. The color is unseen during the fine-tuning process. Since the color of the manipulated object is irrelevant to the language instructions in this case, the robot is expected to behave the same. The results are shown in Table~\ref{tab:CrossEmbodimentTransfer}. As we can see, with little performance drop, the robot is able to generalize to the novel embodiment and the novel scenario.

\begin{table}[!htbp]
    \caption{Generalization on Cup-Placing Task} \centering
    \label{tab:CrossEmbodimentTransfer}
    \begin{tabular}{l c c}
        \toprule
        Scenario & Success Rate & Deviation \\
        \midrule
        Same embodiment, seen color (baseline)     & 0.87 & 0.0287 \\
        Same embodiment, unseen color              & 0.53 & 0.0553 \\
        Novel embodiment, seen color               & 0.84 & 0.0535 \\
        \bottomrule
    \end{tabular}
\end{table}

\subsection{Ablation Study}

In this section, we conduct two ablation studies to investigate the effectiveness of our optimization strategy and the vision-language goal specification. 

\begin{itemize}
    \item We first change the episodic optimization strategy to a fixed-horizon single-iteration optimization. The robot trajectory is optimized for 300 steps with a preset horizon of 100 timesteps. The setting is denoted as \textit{Ours (w/o ep. opt.)}.
    \item We also remove the differentiable rendering module and set up a baseline that uses hand-crafted rewards conditioned on privileged information rather than the rendered visual observation, denoted as \textit{Handcraft (w/o diff. render)}.
\end{itemize}

\begin{table}[!htbp]
    \caption{Ablations on Obstacle-Crossing Task} \centering
    \label{tab:LongHorizonPerformance}
    \begin{tabular}{l c c c}
        \toprule
        Method & Success Rate & Tot. Opt. Steps \\
        \midrule
        Ours                            & 0.72 & 23.9k \\
        Ours (w/o ep. opt.)             & 0.41 & 30k \\
        Handcraft (w/o diff. render)    & 0.19 & 30k \\
        \bottomrule
    \end{tabular}
\end{table}

As shown in Table~\ref{tab:LongHorizonPerformance}, both ablations lead to a significant drop in success rate compared to the original method. With the episodic optimization strategy, our method is capable of handling long trajectory optimization problems, achieving superior performance while consuming fewer optimization steps. 

The integration of differentiable rendering and vision-language models is also crucial for the success of our method. Compared to the baseline using hand-crafted rewards, our method demonstrates the capability of avoiding local minima. This is because the vision-language model can provide more informative feedback associated with task progress. A well-trained vision-language model can guide the robot out of local minima and towards the goal, leading to better performance.

%% file: tex/limit.tex
Though our approach shows promising results in generating robot demonstrations with action annotations, we recognize that its performance is still limited by the quality of the employed vision-language model. Currently, we fine-tune the model using some action-free robotic videos to reduce the domain gap between human data pretraining and robot tasks, as suggested. We believe the need for robotic fine-tuning can be alleviated with a more advanced vision-language model. Our approach can benefit from future advances in representation learning and vision-language pretraining, thus able to be extended to more complex tasks and scenarios.

%% file: tex/con.tex
In this paper, we validate the feasibility of integrating differentiable physics simulation, differentiable rendering, and a vision-language model for automatic and efficient robot demonstration generation in simulation. Given a simulated robot manipulation scenario, our proposed system initiates a sequence of manipulation actions, simulates the manipulation results, and renders the visual observation after manipulation. With a pretrained vision-language model, our system can measure the discrepancy between the observation after manipulation and the text-based task instruction. Thanks to the differentiability of each component, we can directly optimize the action sequence via gradient descent to achieve the instruction. We believe that with the proposed approach to efficiently generating robot demonstrations in simulation and the recent progress in robot simulation task generation, we have the opportunity to scale up robot data for future research.